\documentclass[sensors,article,accept,moreauthors,pdftex,sensors]{Definitions/mdpi} 

\firstpage{1} 
\pubvolume{1}
\issuenum{1}
\articlenumber{0}
\pubyear{2022}
\copyrightyear{2022}
\externaleditor{{Academic Editor: Yi Zhang, Yan Yan and Li Yuan} 
}
\datereceived{29 September 2022} 
\dateaccepted{29 November 2022} 
\datepublished{} 
\hreflink{https://doi.org/} 

\usepackage{amsmath,graphicx}
\usepackage{subfigure}
\usepackage{math}
\usepackage{gensymb}
\usepackage{hhline}
\usepackage{multirow}
\usepackage{color}
\usepackage{marvosym}
\usepackage{algorithm}
\usepackage{algpseudocode}
\usepackage{gensymb}

\Title{Towards Accurate Ground Plane Normal Estimation from~Ego-Motion}

\TitleCitation{Towards Accurate Ground Plane Normal Estimation from Ego-Motion}

\Author{Jiaxin Zhang $^{1}$\orcidA{},
        Wei Sui $^{1}$,
        Qian Zhang $^{1}$,
        Tao Chen $^{2}$\orcidB{} and
        Cong Yang $^{2,*}$\orcidC{}
}

\AuthorNames{Jiaxin Zhang, Wei Sui, Qian Zhang, Tao Chen, Cong Yang}

\AuthorCitation{Zhang, J.; Sui, W.; Zhang, Q.; Chen, T.; Yang, C.}

\address{%
$^{1}$ \quad Horizon Robotics, No. 9, FengHao East Road, Beijing, China; 100094;
             jiaxin02.zhang@horizon.ai (J.Z.); wei.sui@horizon.ai (W.S.); qian01.zhang@horizon.ai (Q.Z.) \\
$^{2}$ \quad School of Future Science and Engineering, Soochow University, Suzhou, China; 215222; chent@suda.edu.cn
}

\corres{Correspondence: cong.yang@suda.edu.cn}

\abstract{In this paper, we introduce a novel approach for ground plane normal estimation of wheeled vehicles. In practice, the ground plane is dynamically changed due to braking and unstable road surface. As a result, the vehicle pose, especially the pitch angle, is oscillating
from subtle to obvious. Thus, estimating ground plane normal is meaningful since it can be encoded to improve the robustness of various autonomous driving tasks (e.g., 3D object detection, road surface reconstruction, and trajectory planning). Our proposed method only uses odometry as input and estimates accurate ground plane normal vectors in real time. Particularly, it fully utilizes the underlying connection between the ego pose odometry (ego-motion) and its nearby ground plane. Built on that, an Invariant Extended Kalman Filter (IEKF) is designed to estimate the normal vector in the sensor's coordinate. Thus, our proposed method is simple yet efficient and supports both camera- and inertial-based odometry algorithms. Its usability and the marked improvement of robustness are validated through multiple experiments on public datasets. For instance, we achieve state-of-the-art accuracy on KITTI dataset with the estimated vector error of 0.39$^\circ$. Our code is available at \href{https://github.com/manymuch/ground_normal_filter}{github.com/manymuch/ground\_normal\_filter}.}


\begin{document}

\section{Introduction}
\label{sec:intro}
\begin{figure}[H]

    \includegraphics[width=0.95\linewidth]{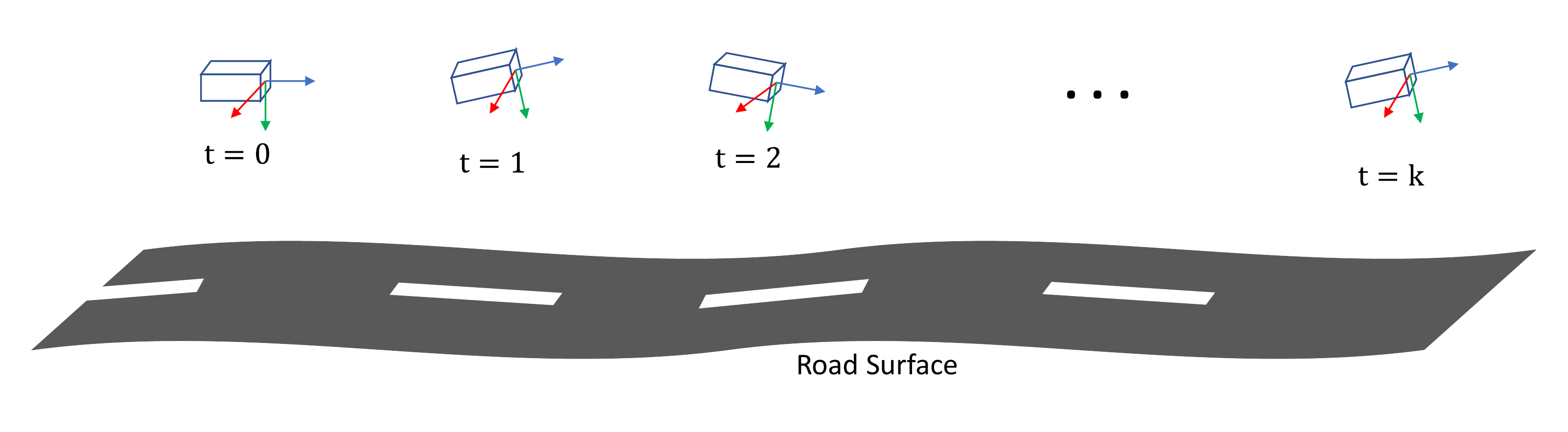}
    \vspace{-0.5em}
  \caption{\small Illustration of a typical dynamic motion of a front-facing camera on a moving vehicle. The pitch angle (rotation around the $x$-axis) is actually oscillating with an amplitude of about 1$^\circ$, though the vehicle moves straight and the road surface looks flat enough. Such pitch angle oscillation is amplified when the vehicle encounters imperfect road surfaces and speed bumps.}
\label{fig:illustration:tisser}

\end{figure}

Accurate ground plane normal estimation is crucial to autonomous driving applications' perception, navigation, and planning. This is because the ground plane in the vehicle's coordinate is dynamically changed due to braking and unstable road surface (see Figure~\ref{fig:illustration:tisser}). As a result, the vehicle pose, especially the pitch angle, oscillates from subtle to obvious~\cite{jazar2008vehicle}. To improve the robustness of autonomous driving system, ground plane normal is estimated and encoded in vision-related tasks, including 3D object tracking~\cite{liu2017adaptive}, lane detection~\cite{wang2004lane,chen2006real,garnett20193d, yang2020towards, qian2019dlt} and road segmentation~\cite{soquet2007road,alvarez2012road, lee2021fast, lee2022joint}, etc. For instance, the ground plane parameters are used for multi-camera calibration in many applications~\cite{knorr2013online, yang2021mlife, yang2022fatigueview}. They are also employed to estimate the depth information of an object on the ground~\cite{liu2007plane, chen2016monocular,qin2022monoground}, and provide vital absolute scale information to the system~\cite{zhou2019ground}. In addition to the aforementioned tasks, existing image-based mapping~\cite{qin2021light} and Bird's-Eye-View (BEV) perception~\cite{reiher2020sim2real,philion2020lift,li2022hdmapnet} algorithms are also sensitive to the accuracy of the ground plane normal parameters. For instance, some BEV-based algorithms are applied with inverse perspective mapping (IPM) with extrinsic parameters from the image plane to the ground plane, thereby mapping pixels from image space to BEV space. 

\begin{figure}[H]

    \includegraphics[width=0.52\linewidth]{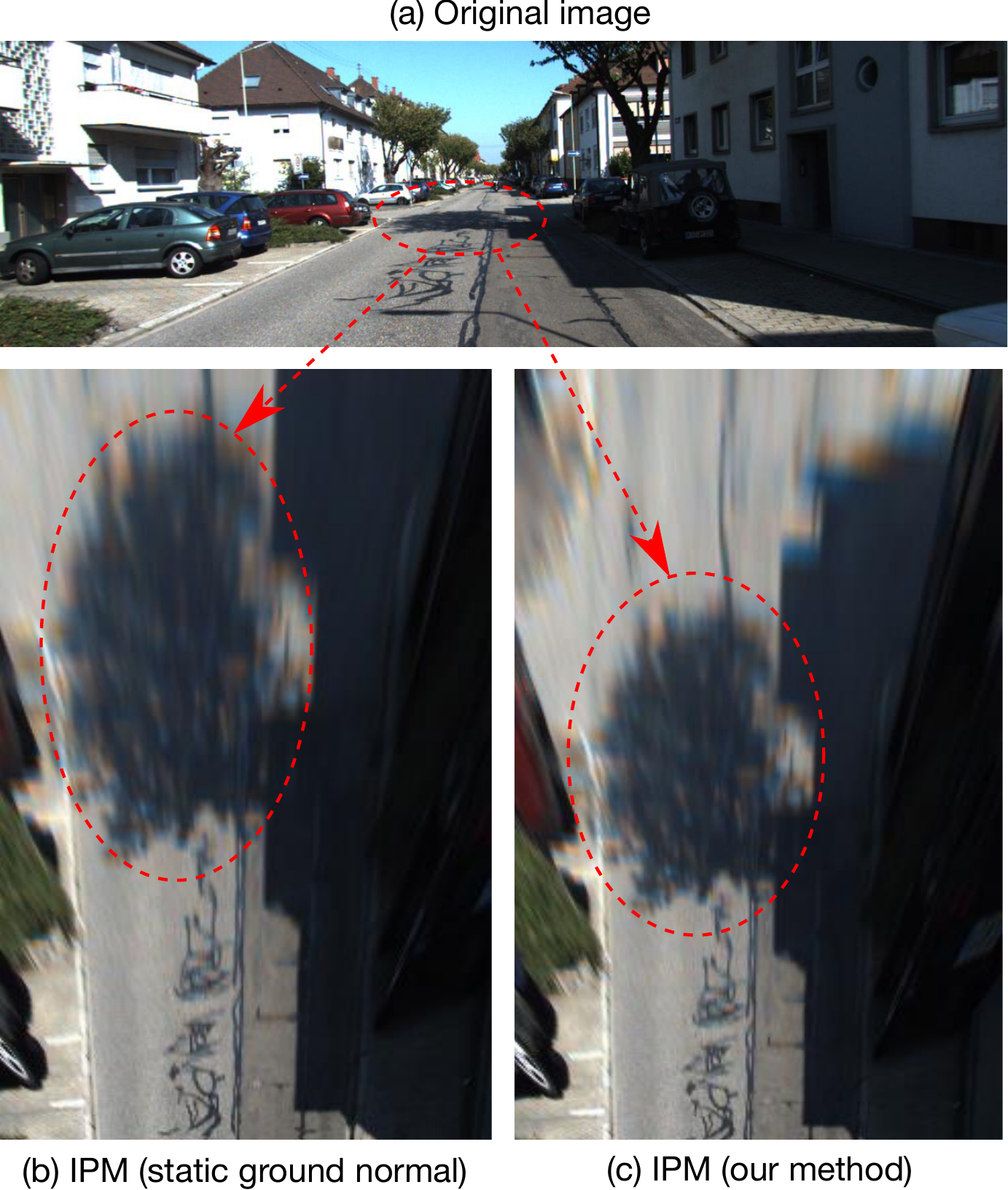}
  \caption{\small Comparison of IPM images before and after using our proposed method. (\textbf{a}) Original image from KITTI odometry dataset. (\textbf{b}) IPM image using fixed extrinsic from the camera to the ground. (\textbf{c}) IPM image using the dynamic extrinsic calculated by our proposed methods. It can be clearly observed that the image in (\textbf{c}) is more accurate. See our supplementary video for better visualization.}
\label{fig:intro:compare}

\end{figure}

However, estimating accurate ground plane normal in real-time is challenging, especially in a monocular setup. The main reason is that the subtle dynamics of the ground plane normal
reflect little in image spaces. Traditional methods usually first estimate homography transform, then decompose it into ground plane normal and ego-motion~\cite{zhou2006robust, dragon2014ground}. Recently, some neural networks were proposed to estimate the depth and normal simultaneously at the pixel level, with photometric and geometric consistency~\cite{sui2021road, xiong2020joint, man2019groundnet}. However, these image-based methods suffer from inadequate accuracy due to a loose connection between the ground plane normal dynamics and image clues. Besides, most previous works simplify (or assume) that the ground plane normal vector of a moving vehicle is constant, which is contrary to the facts. In practice, the normal vector is slightly oscillating when the vehicle moves, even if the road surface seems flat. For instance, a 4-wheel sedan moves along a straight street with a front-facing camera mounted on the top of the windshield. The camera pitch angle (relative to the ground) usually oscillates with an amplitude around 1$^{\circ}$. Though such dynamics reflect little in image spaces, it can be easily observed in the BEV space after image projecting using IPM with fixed extrinsic (see Figure~\ref{fig:intro:compare}a and supplementary video for better visualization).
This phenomenon is also positively verified by our quantitative and qualitative experiments in Section~\ref{sec:experiments}.

We introduce a simple yet efficient method to estimate ground plane normal from ego-motion. Particularly, our approach is compatible with ego-motion provided by SLAM (Simultaneous Localization And Mapping) and SfM (Structure from Motion) algorithms from various sensors (e.g., monocular camera and Inertial Measurement Unit (IMU)). To do so, we design an Invariant Extended Kalman Filter (IEKF) to model the dynamics of the vehicle's ego-motion and estimate the ground plane normal in real-time. Besides, our approach can be easily plugged into most autonomous driving systems that provide ego-motion with little computational cost. As presented in Figure~\ref{fig:intro:compare}, after applying our proposed method, the image quality is dramatically improved. Our experiments in Section~\ref{sec:experiments} verify its effectiveness: the estimated vector error is reduced from 3.02$^\circ$ with~\cite{xiong2020joint} to 0.39$^\circ$ with our proposed method on the KITTI dataset~\cite{Geiger2012CVPR}.

Succinctly, the main contributions of this work are as follows: (1) We introduce a simple yet efficient approach for real-time ground plane normal estimation. (2) The proposed method supports both camera- and inertial-based odometry algorithms thanks to the special design that fully utilizes the ego-motion information as input. Hopefully, our observations and contributions can encourage the community to develop more ground normal estimation methods towards robust autonomous driving systems in the real-world.


\section{Related Works}
\label{sec:related}
We present a concise survey of existing ground normal estimation methods using depth sensors, stereo cameras, and monocular cameras. Some CNN-based methods are also discussed in this part. For a more detailed treatment of this topic in general, the recent compilation by Man and Weng~\cite{man2019groundnet} offers a sufficiently good review.

\subsection{Ground Normal Estimation Using Depth Sensors}
To obtain accurate ground plane parameters, using active depth sensors such as LiDAR and Time of Flight (ToF) is a reliable solution~\cite{gallo2008robust,yu2014obstacle}. While the accurate 3D structure of the environments can be obtained in the form of point clouds (from LiDAR and ToF), ground plane parameters can be easily estimated by plane fitting. Built on that, the least square method is employed once the points belong to the ground~\cite{choi2014robust,zhang2010lidar}. For application, existing LiDAR-based works only are triggered to estimate ground planes in some challenging scenarios, such as offroad~\cite{mcdaniel2010ground} and construction areas~\cite{mikadlicki2017ground}. However, our proposed method takes ego-motion as input and can be easily plugged into most autonomous driving systems. As a result, our method is more general and can be employed in most driving scenes.

\subsection{Ground Normal Estimation Using Stereo Cameras}
Cheaper than active depth sensors such as LiDAR and ToF, stereo cameras are more accessible and can provide reasonable depth information through disparity. Similarly, most stereo-based methods are also designed to deal with particular cases, such as staircase~\cite{lee2012real} and cluttered urban environments~\cite{schwarze2015robust}. However, they normally require good lighting conditions and rich textures. While depth and normals are highly related to 3D information, they are jointly trained with stereo images and consistency loss~\cite{kusupati2020normal}. To directly model road surface with a plane normal, Stephen {et al.} estimate the ground plane based on disparity, thereby detecting and tracking obstacles and curbs~\cite{se2002ground}. Nikolay {et al.} also propose to use dense stereo disparity for ground plane normal estimation~\cite{chumerin2008ground}. These disparity-based methods usually focus on analyzing the ground plane together with objects and the 3D structure of the road in detail. In comparison, our approach only requires a monocular camera or even IMU-only odometry to obtain high-accuracy ground plane normals in real~time.

\subsection{Ground Normal Estimation Using Monocular Camera}
Ground plane estimation from a monocular camera is challenging, as it attempts to reason 3D information from 2D images. The connection between ground plane normals and ego-motion is initially modelled in HMM (Hidden Markov Model)~\cite{dragon2014ground}, then the odometry and ground plane normals are jointly estimated from image sequences. Zhou {et al.} propose to use constrained homography to estimate the ground plane for robot platforms~\cite{zhou2006robust}. In terms of monocular Visual Odometry (VO), it is common to combine the ground plane estimation with scale recovery~\cite{song2014robust,zhou2016reliable}. Our method is fundamentally different from the aforementioned methods: we decouple those multiple tasks and only estimate the normal vector from ego-motion with our specially designed IEKF. In such a way, our proposed method supports monocular setup and other algorithms with different sensors, such as IMU-only odometry.

Recently, some Convolutional Neural Networks (CNN) have also been proposed to estimate ground planes. Particularly, given a monocular image sequence, photometric consistency can be used with homography warping to recover the normal vector in a self-supervised manner~\cite{sui2021road,xiong2020joint}. To further improve the accuracy, GroundNet~\cite{man2019groundnet} jointly learns pixel-level normals, ground segmentation, and depth maps using multiple networks. As a result, their latency is relatively high, ranging from 130 to 920 milliseconds/frame. While our method mainly focuses on the ground plane normal vector estimation using ego-motion, as detailed in Section~\ref{sec:experiments}, the latency is reduced to 3 milliseconds/frame for the IMU odometry and 50 milliseconds/frame for the monocular visual odometry. Moreover, the ground truth of ground plane normal is difficult to obtain and verify. Most existing works apply homography transformation on original images to produce augmented inputs and corresponding labels~\cite{man2019groundnet,dragon2014ground}. These methods consider normal vectors of original images as the fixed value calculated by extrinsic. However, in practice, such an approximation is inaccurate, and the augmentation deviates the data distribution from actual use cases. Instead, we use LiDAR points to calculate the ground plane normal as ground truth. The effectiveness is qualitatively and quantitatively verified in Section~\ref{sec:experiments}.

\section{Ground Plane Normal}
\label{sec:dynamic}
In this part, we structurally study the dynamics of the ground plane normal, thereby verifying the motivation behind this work. We argue that the ground plane normal vector in a vehicle's reference system is
oscillating when the vehicle is moving. To verify it, we take a clip from KITTI~\cite{Geiger2012CVPR} odometry sequence \# 00 for illustration. Theoretically, if the ground plane normal remains constant, the IPM images (with fixed extrinsic) should be similar (e.g., the parallel road lanes and edges) between adjacent frames.

\begin{figure}[H]

    \includegraphics[width=0.77\linewidth]{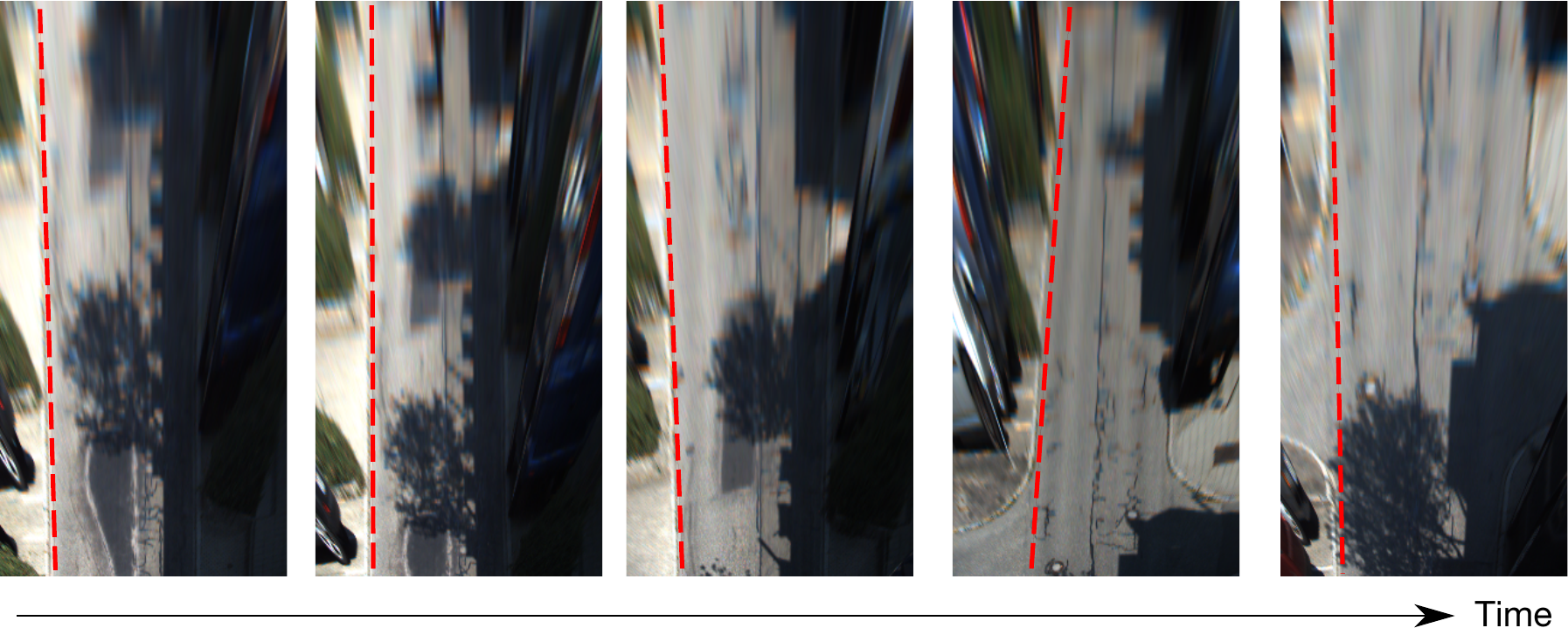}
  \caption{\small IPM images with the constant ground plane normal: road edges are not properly aligned.}
\label{fig:normal_constant}

\end{figure}

\begin{figure}[H]

    \includegraphics[width=1\linewidth]{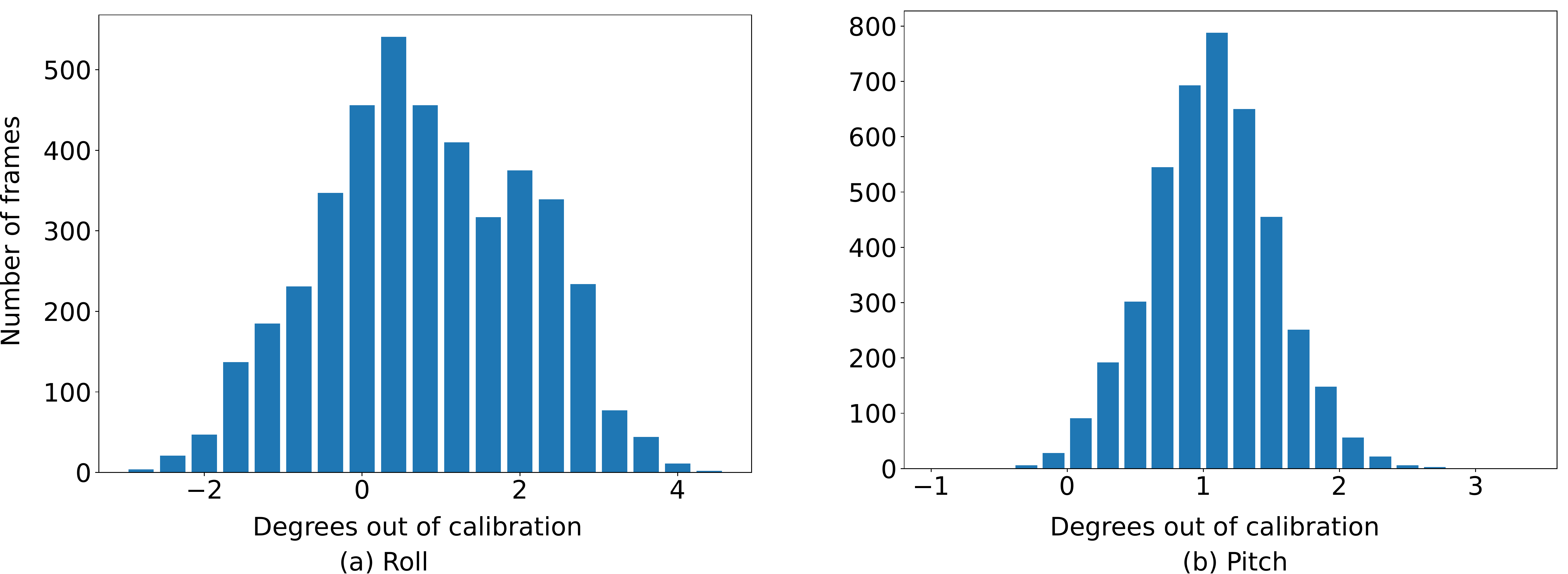}

  \caption{\small Statistics of frames (KITTI odometry sequence \# 00) that are out of calibration in pitch and roll.}
\label{fig:normal_dynamic}

\end{figure}

However, as visualized in Figure~\ref{fig:normal_constant}, the road edges between adjacent frames (with fixed extrinsic) are not well aligned after IPM with a constant ground plane normal. To explore this phenomenon, we use LiDAR points from the dataset to calculate the ground truth (GT) of the ground plane normal. Built on that, the GT road edges are marked in red dot lines. We clearly find that most real road edges are not properly aligned with GT, with more than 1$^\circ$ out of calibration. To get more general statistics of such dynamics, we count the number of frames based on their variation to the GT in roll and pitch. The final statistics are presented in Figure~\ref{fig:normal_dynamic}. It can be observed that the mean variation of pitch and roll angles are around $1.2^\circ$ and $3.5^\circ$, respectively. In other words, rather than constant, the ground plane normal vector is dynamically changed when the vehicle is moving.

\vspace{-6pt}

Similarly, Table~\ref{table:kitti_dynamics} presents the mean values of pitch and roll dynamics on all KITTI odometry sequences. We can draw the same conclusion: the ground plane normal is not constant (around $1^\circ$) when a vehicle is moving. Such instability could further influence the performance of autonomous driving tasks. Therefore, our estimated ground plane normal vector (in Section~\ref{sec:approach}) can be encoded to improve the robustness of autonomous driving applications. 

\begin{table}[H]

\caption{Statistics of pitch and roll dynamics on all KITTI odometry sequences.}
\label{table:kitti_dynamics}

\begin{adjustwidth}{-\extralength}{0cm}
\newcolumntype{C}{>{\centering\arraybackslash}X}
\begin{tabularx}{\fulllength}{lCCCCCCCCCCCCC}

   \toprule
   	\textbf{Sequence} & \textbf{00}   & \textbf{01}   & \textbf{02}   & \textbf{03}   & \textbf{04}   & \textbf{05}   & \textbf{06}   & \textbf{07}   & \textbf{08}   & \textbf{09} & \textbf{Mean} & \textbf{Std} \\ \midrule
	Pitch       & 1.06 & 1.16 & 1.11 & 0.40 & 1.21 & 1.27 & 1.27 & 1.27 & 1.31 & 1.47 & 1.15 & 0.27\\
	Roll        & 0.92 & 0.59 & 1.20 & 1.30 & 1.46 & 0.99 & 0.78 & 0.70 & 0.93 & 0.91 & 0.98& 0.26 \\ \bottomrule
  \end{tabularx}
\end{adjustwidth}
\end{table}

\section{Approach}
\label{sec:approach}
\begin{figure}[H]

    \includegraphics[width=1\linewidth]{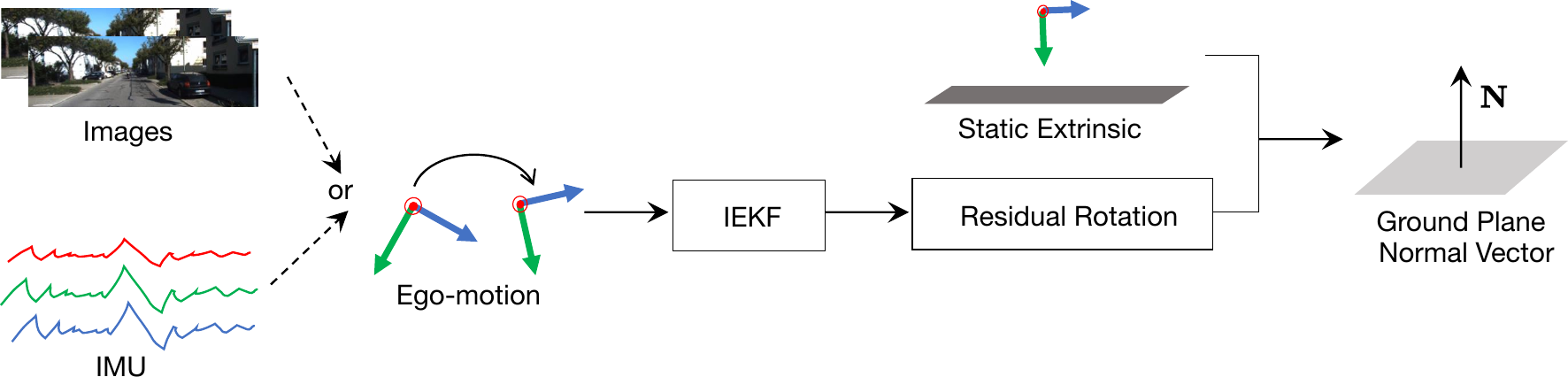}
  \caption{\small Overview of our proposed ground plane normal estimation pipeline. Our proposed IEKF can process ego-motion from various sensors, such as IMU, visual odometry from monocular images, and SLAM systems that can provide real-time odometry information. The final ground plane normal vector $N$ is predicted in real-time based on the combination of residual rotation from IEKF and static extrinsic from prior calibration.}
\label{fig:pipeline}

\end{figure}

In this part, our proposed ground plane normal estimation method is detailed. Figure~\ref{fig:pipeline} presents the pipeline. In short, we formulate the relationship between the odometry (from images or IMU) and ground plane normal based on IEKF. For a better description, We use a front-facing monocular camera on a wheeled vehicle as an example in the following descriptions.

For a moving vehicle, its camera pose is trivially coherent with the ground plane. In real environments, the road surface is not ideally plane, but a segment close to the camera is approximately flat. In such a case, it is applicable to calculate the normal vector of the segment in the camera reference system. Specifically, when the vehicle is static, the ground plane normal vector can be computed from extrinsic parameters between the camera and the ground plane. The extrinsic can be easily obtained via off-line checkboard calibration~\cite{hartley2003multiple}. When the vehicle is moving, due to oscillations of roll and pitch angles (see Figure~\ref{fig:illustration:tisser}), the extrinsic is no longer accurate to represent the relationship between the camera and the ground plane. In such a case, our proposed method is triggered. The rationale behind our method is that the dynamics of a normal vector can be roughly divided into two parts in the frequency domain: The low-frequency part describes the actual elevation changes, such as bumps and bridges. The high-frequency part is the oscillation, mainly because of braking and acceleration. Our goal is to split these two components from ego-motion to calculate the ground plane normal vectors. In summary, our proposed method is built on two assumptions: (1) The close-to-camera road surface can be approximated as a flat plane. (2) The mean camera pose is close to its static extrinsic calibration.

\begin{figure}[H]

    \includegraphics[width=0.6\linewidth]{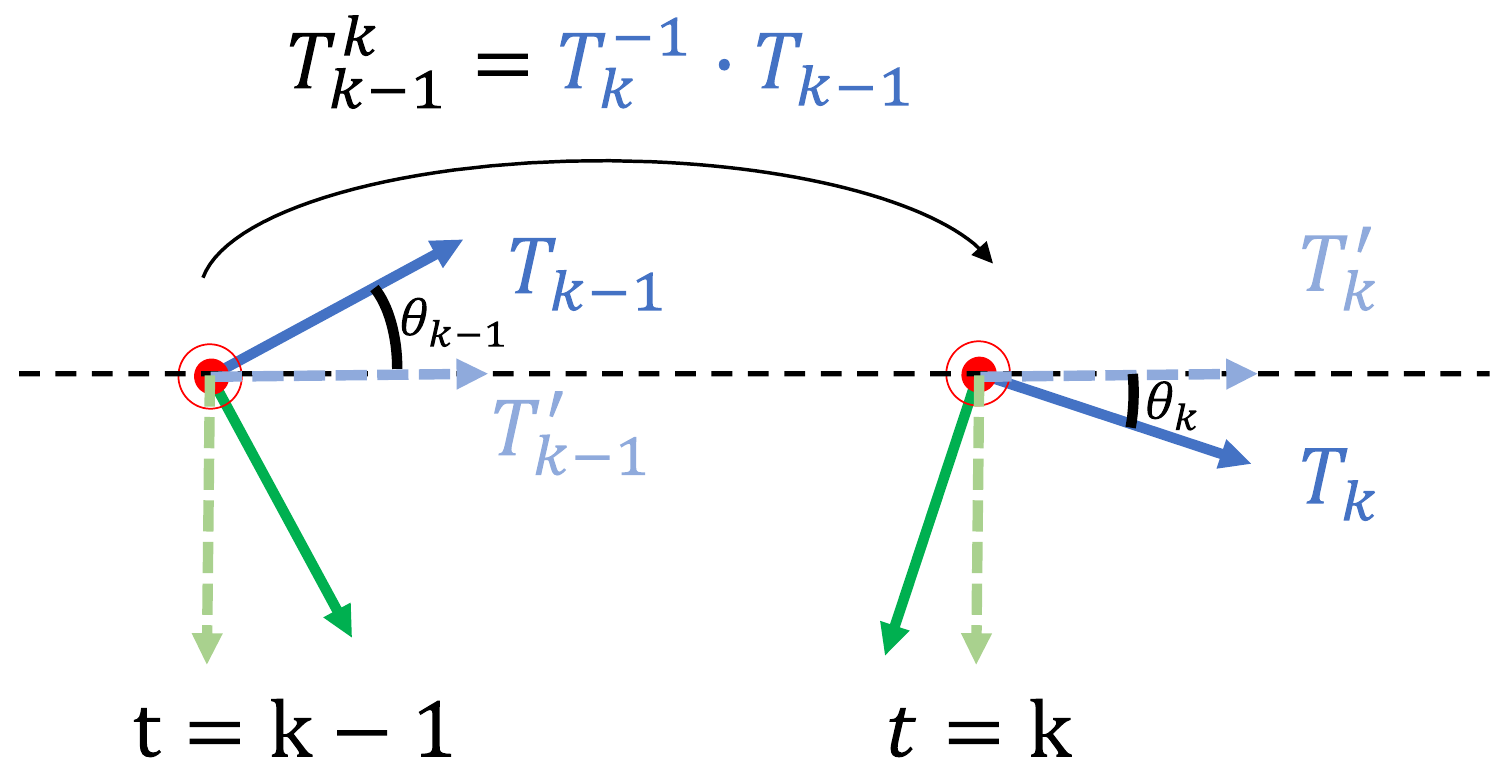}
  \caption{\small 2D side view of the camera reference system in two adjacent frames. $T_{k-1}^{'}$ and $T_{k}^{'}$ are the ideal camera reference system when the vehicle is stopped. $T_{k-1}$ and $T_{k}$ are the actual camera poses. $T_{k-1}^k = T_{k}^{-1} \cdot T_{k-1}$ is the ego-motion between two frames. The black dashed line is the ideal horizontal line parallel to the ground plane. $\theta_{k-1}$ and $\theta_{k}$ are the pitch angles relative to the ground plane. The actual camera extrinsics to the ground plane are $T_{k-1} \cdot T_{k-1}^{'}$ and $T_{k} \cdot T_{k}^{'}$, which is equivalent to the ground plane normal vector. Best view in colour.}
\label{fig:detail}
\end{figure}

Figure~\ref{fig:detail} presents the camera reference system of two adjacent frames. The transformation between the actual (vehicle moving, dark green and blue) and ideal (vehicle stopped, light green and blue) camera reference system is equal to the extrinsic rotation between the camera and ground plane. Accordingly, it can be used to calculate the ground plane normal vector:
\begin{equation}
\small
	\label{eqn:1}
	N_k = \mathcal{N}(T_{k}^{-1} \cdot T_{k}^{'})
\end{equation}
where $\mathcal{N}(\cdot)$ means extracting the second column (y-component) of the rotation from a transformation matrix. The rotation of $T_{k} \cdot T_{k}^{'}$ can be decomposed to Euler angles: roll (z-axis), pitch (x-axis) and yaw (y-axis). For a moving vehicle, as shown in Figure~\ref{fig:normal_dynamic}, the pitch angle is the most dynamic component. Our task is to estimate pitch angle ($\theta_k$ in Figure~\ref{fig:detail}), or more generally the residual rotation $T_{k} \cdot T_{k}^{'}$:
\begin{equation}
\small
	T_{k-1}^k = T_{k}^{-1} \cdot T_{k-1}.
\end{equation}

As $T_{k}$ is available from ego-motion, the problem is now turned into estimating and tracking the ideal (vehicle stop) camera reference system $T_{k}^{'}$. At first glance, this is trivial since $T_{k}^{'}$ is static and always parallel to the ground. However, the only input is ego-motion (the transformation between adjacent frames in the world reference system (WRS)). Even if the WRS is
aligned with the ground plane, ego-motion unavoidably suffers from the drifting of long sequences. Thus, estimating $T_{k}^{'}$ from limit frames of history odometry is necessary, which intuitively leads to Kalman filter~\cite{kalman1960new} as a potential solution. 

To do so, we adopt the idea of IEKF~\cite{bonnabel2007left,barrau2016invariant} to our rotation estimation scenario. The general idea of IEKF is to use a deterministic non-linear observer directly on Lie groups instead of using a correction term on linear output. As shown in Figure~\ref{fig:pipeline}, our method takes ego-motion as the input and output ground plane normal vector $\textbf{N}$. The source of ego-motion can be chosen from monocular SLAM system~\cite{mur2015orb,engel2017direct}, learning-based monocular odometry~\cite{yang2020d3vo,zhang2021deep,wagstaff2022coupling,zhang2022towards}, pure IMU-based odometry~\cite{brossard2020ai}, and other SLAM (or odometry) algorithms that provide real-time ego-motion between frames.

The whole procedure of our proposed method is described in Algorithm 1. In terms of IEKF, it is adopted as follows: The state of the filter is a member of SO(3), as we only consider the rotation of the sensor. The state and its covariance are initialized with zeros and identity matrix, respectively. We only consider a zero-order state (SO(3)), i.e., the process model is an identity function on the input rotation. Higher order state (e.g., angular velocity) could be added to the filter if the source odometry sensor provides such observation, such as IMU. Nevertheless, we found that a constant process model is enough in our cases and makes our approach more general. If the sensor odometry provides relative transformation (i.e., $T_{k}^{k-1}$), the absolute transformation (i.e., $T_k$) is tracked by integration over time. The observation of the filter is the rotation part of $T_k$. To calculate the normal vector ($N_i$) of the current frame, the residual rotation ($G_i$) is calculated by the difference between the prediction state ($Y_i$) and absolute transformation ($T_k$). Note that the prediction state is calculated before the observation of the current frame is applied to the filter.

\begin{algorithm}
	\caption{Ground Plane Normal Vector Estimation}
	\begin{flushleft}
		\textbf{Require:} Extrinsic calibration between reference sensor and ground plane $E_{r}^{g}$\\
		\textbf{Input:} Ego-motion from the reference sensor: [$T_0, T_{1}^{0}, ..., T_k^{k-1}$]\\
		\textbf{Output:} Ground plane normal vector w.r.t reference sensor: [$N_1, N_2, ...,N_k$]
	\end{flushleft}
	\begin{algorithmic}
		\State \textbf{Initialization:} \\
		Covariance matrix $C = I_{3}$ \\
		Initial state $x \in SO(3)$ \\
		Process model $x = f(x)$\\
		Process variance $P = p \cdot I_{3}$ \\
		Measurement model $\hat{x} = x $ \\
		Measurement variance $M = I_{3}$ \\
		Invariant Extended Kalman Filter $\mathcal{F}(x, C, P, M)$ \\
		Cumulative ego odometry $T_i \in SO(3)$
		\For {$t=0,1, \ldots,k$}
		\State Compute $T_t = T_{t-1} \cdot T_{t}^{t-1}$
		\State Predict state: $T^{'}_t = \mathcal{F}.predict()$
		\State Update filter: $\mathcal{F}.update(T_t)$
		\State Compute residual rotation: $G_t = T_t^{-1} \times T^{'}_{t} $
		\State Compute normal vector $N_t$ from residual rotation $G_t$ using Equation \ref{eqn:1}
		\EndFor
	\end{algorithmic}
\end{algorithm}

\section{Experiments}
\label{sec:experiments}
In this part, we first introduce the implementation details of our proposed method. Built on that, we evaluate its performance quantitatively and qualitatively. Finally, the limitations of our method are discussed.

\subsection{Implementation}
\label{sec:experiments:imp}
To validate that the proposed method is agnostic to the source of ego-motion, we choose two challenging sensor setups for evaluation: monocular camera and pure IMU odometry. The experiments are conducted on the popular KITTI dataset~\cite{Geiger2012CVPR}. It provides four front-facing camera images, raw IMU measurement data, LiDAR points, extrinsic calibration, and ground truth ego-motion. For monocular setup, ORB-SLAM2~\cite{mur2015orb} is applied on the left RGB camera images to obtain ego-motion. In terms of IMU-only odometry, the AI-IMU~\cite{brossard2020ai} is employed to extract ego-motion. After that, the extrinsic is used to convert the ego-motion from the IMU reference system to the camera reference. Note that KITTI provides IMU data at 100 Hz while the camera is running at 10 Hz. To fairly compare different odometry sources, the frame rate of IMU odometry is down-sampled to 10 Hz via integration.

To quantitatively evaluate our proposed method, the ground truth of the ground plane normal is calculated using LiDAR point clouds. Specifically, for each frame, the point cloud is projected onto the image to get 2D-3D correspondence, thereby selecting points within the camera's visual hull. Then, an off-the-shelf semantic segmentation method~\cite{cheng2022masked} is applied to images to obtain image masks for ground areas. Finally, the RANSAC~\cite{fischler1981random} plane fitting is applied on the points which only correspond to the image ground area. For IEKF, the scale of process variance $p$ is set to $10^{-2}$. All our experiments run on a desktop with an Intel i5-6600K CPU running at 3.50GHz. The operating system is Ubuntu 18.04.6 LTS. Note that, unlike GroundNet~\cite{man2019groundnet}, GPU is not required by our proposed method.

\subsection{Quantitative Evaluation}
Here, the estimated ground plane normal vectors are evaluated against ground truth:
\begin{equation}
\small
	 E_{rad} = \frac{\sum_{i = 1}^{k}\arccos{(\vec{N^{est}_{i}} \cdot \vec{N^{gt}_{i}})}}{k},
\end{equation}
where $E_{rad}$ is vector errors in radians, $\vec{N^{est}_{i}}$ and $\vec{N^{gt}_{i}}$ are the estimated and ground truth vectors in $i-th$ frame, respectively. All the normal vectors are unitary vectors with modulus = 1. As mentioned in Section~\ref{sec:experiments:imp}, there are two types of ground truth: fixed extrinsic and plane fitting. For the first one, the ground truth normal vector is constant and calculated from calibration. For the second one, the ground truth normal vectors are calculated by the plane fitting from LiDAR points. For a fair comparison, we keep the original setting of existing methods and apply our method to both IMU and monocular sensors. Table~\ref{tab:comparison} presents the detailed results. As presented in Table~\ref{tab:comparison}, our proposed method achieves the best accuracy in both sensors. For instance, the estimated vector error on the KITTI dataset is reduced from 3.02$^\circ$ with~\cite{xiong2020joint} to 0.39$^\circ$ with our proposed method. Moreover, the monocular-based method provides slightly better results compared with IMU-only odometry. This is because the accuracy of monocular odometry is inherently higher than IMU odometry. We also compare the computation time (if provided) in Table~\ref{tab:comparison}. It can be clearly observed that the computation time with our method is between 3--50 ms per frame, dramatically reduced as well. Overall, our proposed method can estimate accurate ground plane normal vectors in real time.

\begin{table}[H]
\caption{Quantitative comparison of our proposed method with previous works. The running time is also compared here to demonstrate the improvement in efficiency using our method. Particularly, the adopted IEKF takes less than one millisecond per frame.}
\label{tab:comparison}

\begin{adjustwidth}{-\extralength}{0cm}
\newcolumntype{C}{>{\centering\arraybackslash}X}
\begin{tabularx}{\fulllength}{lCCC}

		\toprule
		\textbf{Methods}                              & \textbf{Error (\boldmath{\degree})} & \textbf{Time (ms/frame)} \\ \midrule
		HMM~\cite{dragon2014ground}          & 4.10            & -             \\
		Xiong~\cite{xiong2020joint}   & 3.02            & -             \\
		GroundNet~\cite{man2019groundnet}    & 0.70            & 920             \\
		Road Aware~\cite{sui2021road}        & 1.12            & 130             \\
		Naive~\cite{fischler1981random}      & 0.98            & -             \\
		Ours (IMU)                           & 0.44            & 3 = 2 (IMU odometry) + 1 (IEKF)             \\
		Ours (Monocular)                     & 0.39            & 50 = 49 (Visual odometry) + 1 (IEKF)     \\ \bottomrule
	\end{tabularx}
\end{adjustwidth}
\end{table}

\subsection{Qualitative Evaluation}
\begin{figure}[H]

    \includegraphics[width=0.65\linewidth]{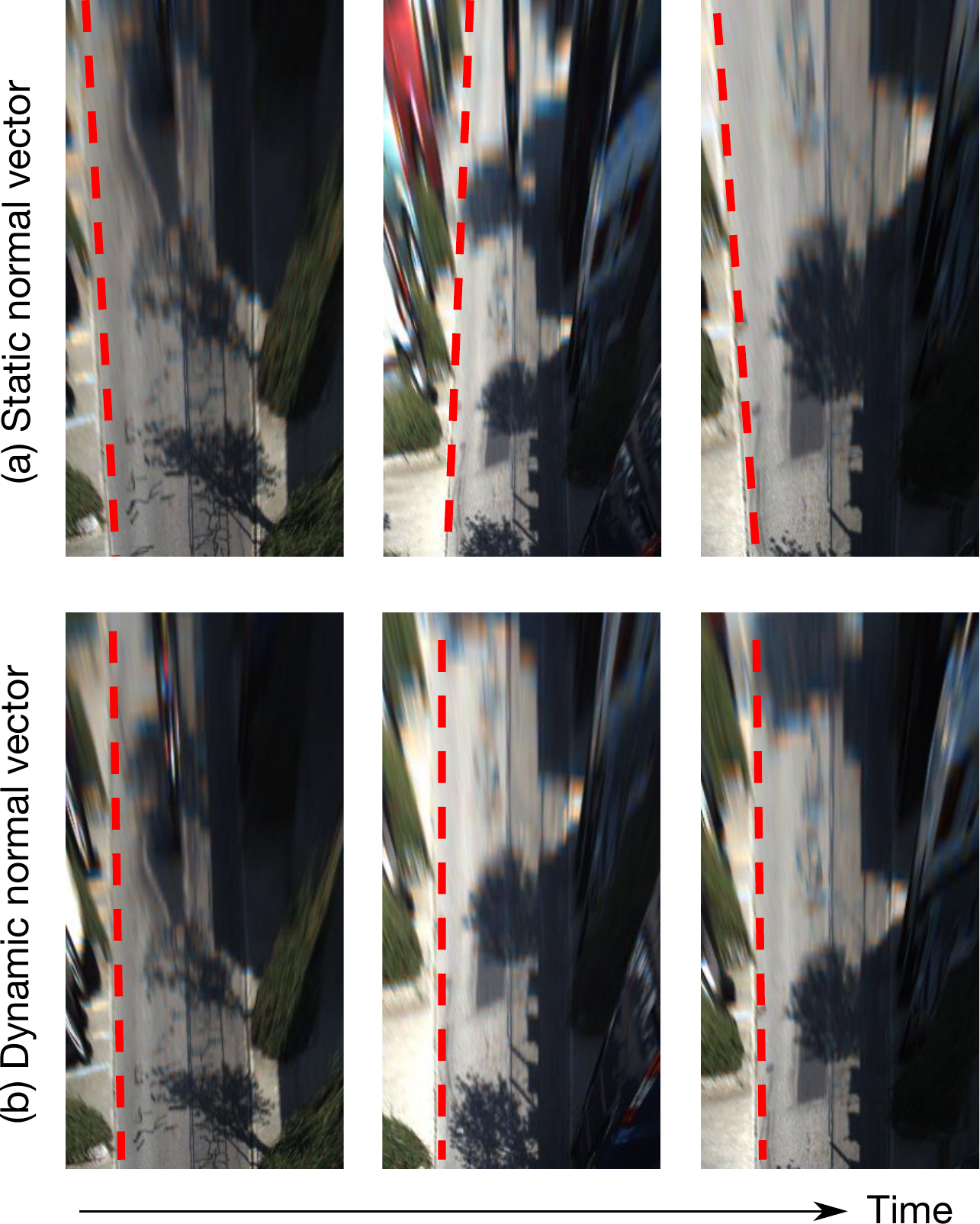}
  \caption{\small Visual comparison of IPM images using (\textbf{a}) static normal vector based on fixed extrinsic calibration, and (\textbf{b}) dynamic normal vector using our proposed method. The odometry input is formed by the monocular version of ORB-SLAM2. We can clearly find that the road edges are not parallel with each other with a static normal vector. Based on the dynamic normal vector from our method, the road edges in IPM images are more parallel and consistent.}
\label{fig:visualization}

\end{figure}

To better understand our contributions, the IPM images with static (from fixed extrinsic calibration) and dynamic (from our proposed method) normal vectors are visually compared in Figure~\ref{fig:visualization}. Here, static normal vector means the ground plane normals are constant~\cite{xiong2020joint}. Ideally, if the ground plane normals used in IPM are accurate, the parallel lanes on the flat road surface should maintain parallel in IPM images (see Section~\ref{sec:intro}). However, as shown in Figure~\ref{fig:visualization}a, the road lanes are not properly parallel with the static normal vector. However, with dynamic normal vector from our method, the road edges in IPM images are more parallel and consistent in Figure~\ref{fig:visualization}b.

\begin{figure}[H]

    \includegraphics[width=0.7\linewidth]{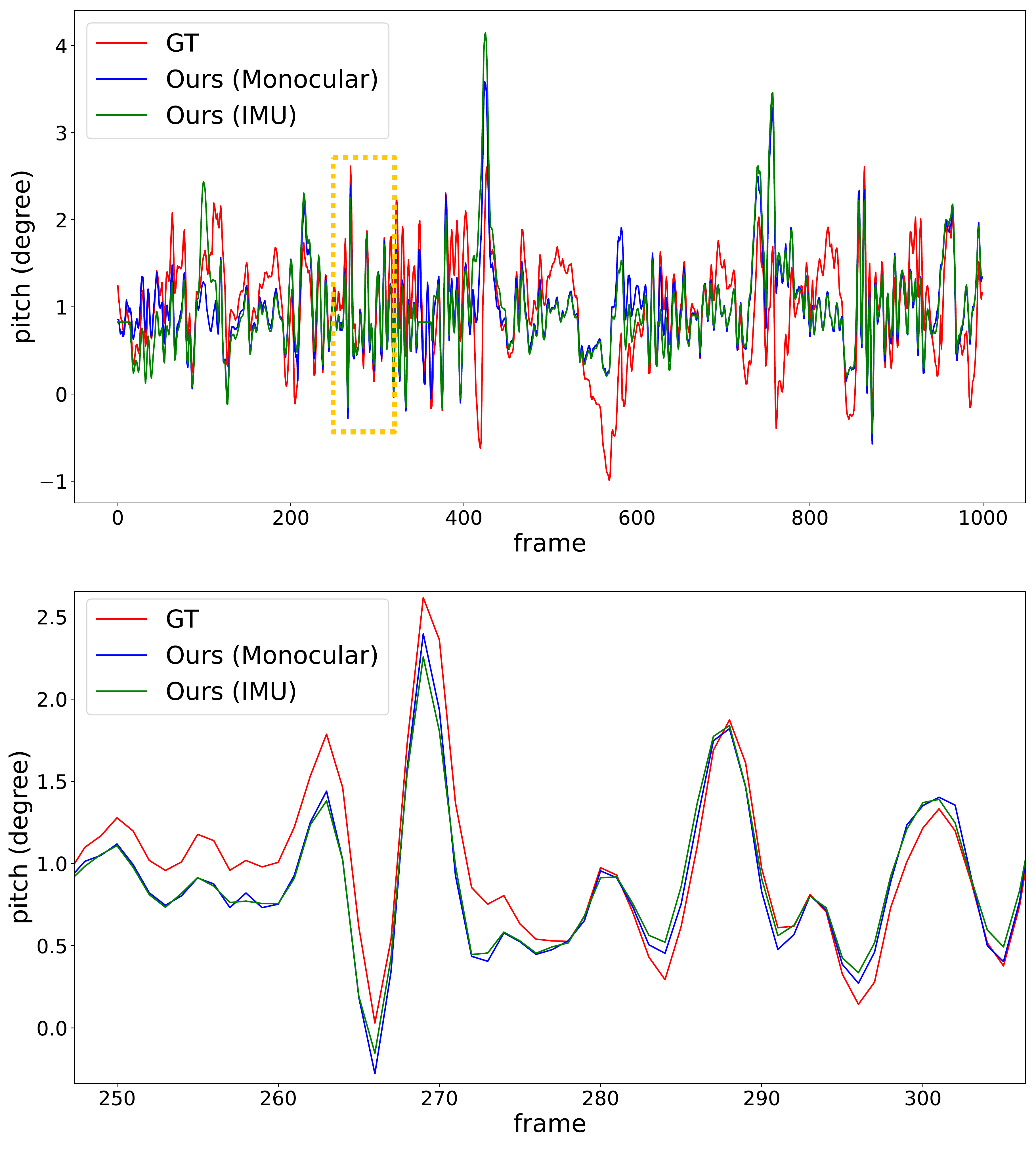}
  \caption{\small Plots of pitch angles with normal vectors calculated by the proposed methods. The bottom plot shows the details within 50 frames from the orange box. The oscillation tendency of the pitch angles from the proposed methods aligns well with the ground truth. Note that the overall amplitude of the pitch angles is actually small, usually within 1 degree.}
\label{fig:pitchplot}

\end{figure}

\begin{figure}[H]

    \includegraphics[width=0.7\linewidth]{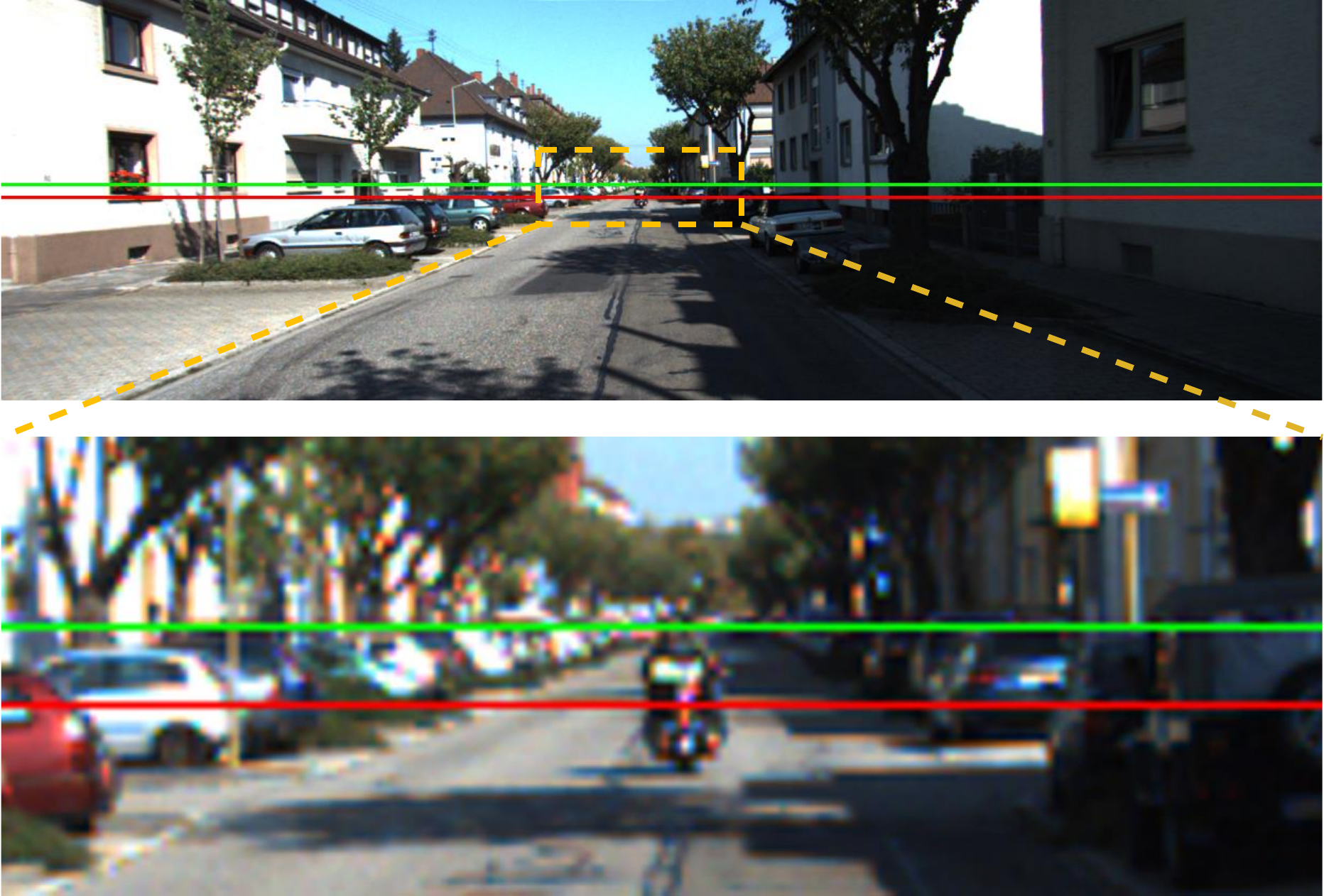}
  \caption{\small Visualization of vanishing lines. The red and green horizontal lines are vanishing lines converted from fixed and dynamic ground plane normals, respectively. The \textbf{bottom} image is a zoom-in image of orange rectangular areas from the \textbf{top} image. The green line is obviously a more accurate estimation of the vanishing line.}
\label{fig:vanishing}

\end{figure}

Figure~\ref{fig:pitchplot} details pitch angle variations in the IPM sequence of the KITTI odometry sequence-00 clip. Based on the dynamic normal vectors using our proposed method, we can clearly find that the pitch angles (both monocular camera and IMU) are properly aligned with ground truth among most frames. However, in some cases (frame 500 to 600), the estimated ground normals differ from the GT. The reason is that the vehicle is making a sharp right turn, and the proposed method with IEKF can not produce an ideal estimation under extreme vehicle dynamics.
As discussed in~\cite{wang2004lane,chen2006real,garnett20193d}, normal vector estimation is inherently equivalent to vanishing lines estimation. Thus, converting ground normals into vanishing lines (in original image space) can also provide convincing visualization of our proposed method. In Figure~\ref{fig:vanishing}, the green line is calculated from our proposed method, showing a reasonable vanishing line estimation. The red line is calculated from static calibration (static normal vector) and apparently deviates from the ideal one. A better visualization can be found in the supplementary video.

\vspace{-6pt}

\begin{figure}[H]

    \includegraphics[width=0.9\linewidth]{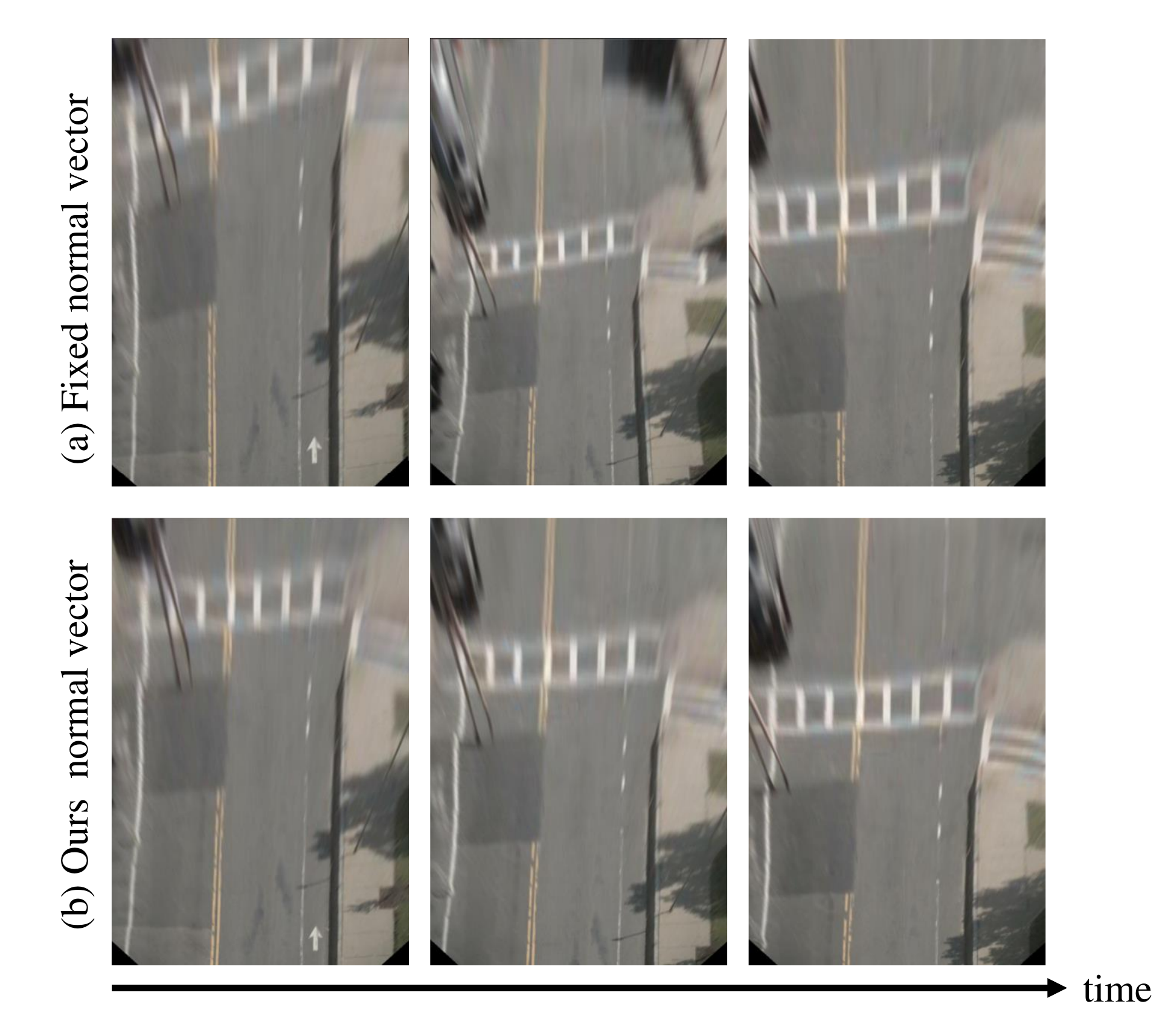}
  \caption{\small IPM visualization on the nuScenes dataset.}
\label{fig:nuscenes}

\end{figure}

To verify the robustness of our proposed method, we conduct the same experiments on the nuScenes~\cite{nuscenes} dataset. As shown in Figure \ref{fig:nuscenes}, the images on the left are IPM results using origin fixed camera extrinsic. The images on the right show IPM results using ground plane normals estimated by our proposed methods. We can clearly see that the proposed method produces more stable and reasonable IPM images.

\subsection{Ablation Study}
\begin{figure}[H]

    \includegraphics[width=1.0\linewidth]{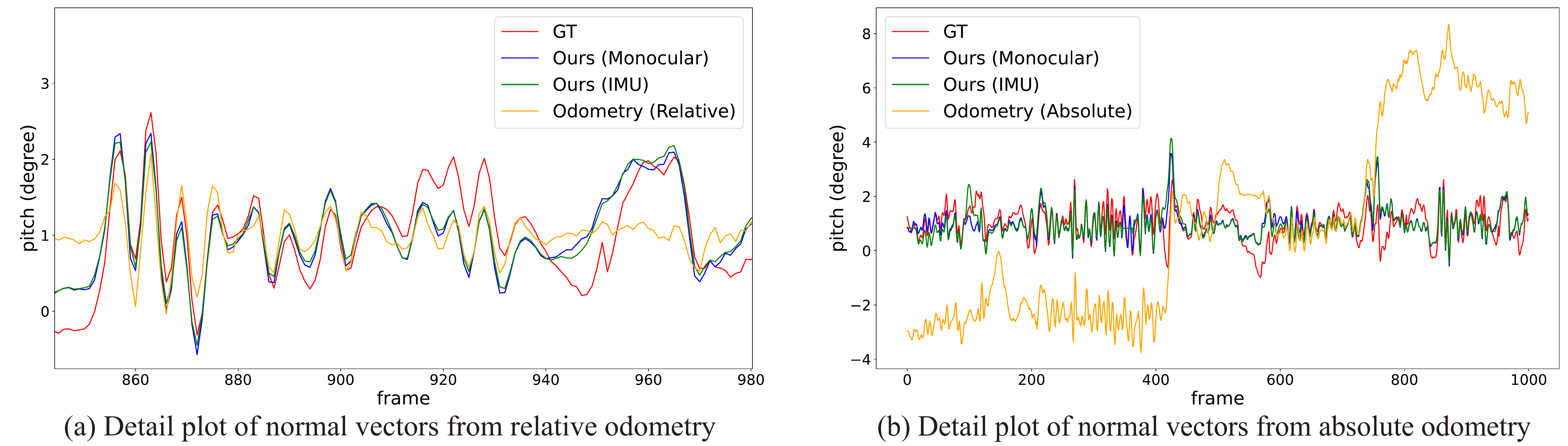}
  \caption{\small Comparing ground plane normal estimated by odometry only.}
\label{fig:odometry}

\end{figure}

To evaluate the effectiveness of using IEKF on the odometry to calculate the ground plane normal, we conduct extra experiments by using odometry only to obtain the ground plane normal. There are two ways to use odometry information directly: relative odometry and absolute odometry. The former is the relative pose between adjacent frames provided by the odometry algorithm, and the latter is accumulated odometry, i.e., current pose w.r.t first frame. As shown in the Figure \ref{fig:odometry}, using pure relative odometry results in inconsistent ground normal estimation in some cases. This is because relative rotation between frames only contains ``instant information'' of the vehicle poses, thus being unable to handle various road surfaces such as small slopes or bumps. For absolute odometry, the result is even worse as it suffers from drifting issues as errors of odometry accumulate over time. Quantitative results are shown in Table \ref{table:ablation}

\begin{table}[H]
\caption{Quantitative comparison ground plane normal estimation between our proposed methods and using odometry directly. }
\label{table:ablation}
\newcolumntype{C}{>{\centering\arraybackslash}X}
\begin{tabularx}{\textwidth}{lCC}

		\toprule
		\textbf{Methods}                              & \textbf{Error (}\boldmath{$^{\circ}$}\textbf{)} \\ \midrule
		Pure odometry(relative)              & 1.09            \\
		Pure odometry(absolute)              & 2.98            \\
		Naive(constant normal)               & 0.98            \\
		Ours(IMU)                            & 0.44            \\
		Ours(Monocular)                      & 0.39            \\ \bottomrule
	\end{tabularx}
\end{table}

\section{Limitations}
\label{sec:limitations}
Though our proposed method can estimate accurate ground plane normal vectors in real-time, there are still two limitations: (1) Our proposed method can only be applied in wheeled vehicles since it relies on the underlying connection between the ground plane and ego-motion in wheeled vehicles. (2) Our proposed method relies on the assumption that the nearby ground plane can always be approximated as a flat plane and the vehicle is driving smoothly. Thus, the estimation accuracy would degrade if the vehicle is driving on extremely uneven roads such as terrains and slopes or making harsh turns. For these cases, the effective range of the ground plane normal estimated by our proposed method could narrow down to smaller areas.

\section{Conclusions}
\label{sec:conclusions}
In this paper, we propose a ground plane normal vector estimation in driving scenes. We structurally study the dynamics of normal vectors when the vehicle is moving, which were previously considered constants. The argument is verified with both visualization and quantitative experiments. After analyzing the underlying connection between ground plane normals and vehicle odometry, the invariant extended Kalman filter is adopted to estimate the normal vectors with high accuracy in real time. The input of the filter is agnostic to the sensors that produce odometry information. Experiments on public datasets demonstrate that our method achieves promising accuracy on both monocular and IMU-only odometry.

\vspace{6pt} 


 \supplementary{The
 following are available at \linksupplementary{s1}, Figure S1: title, Table S1: title, Video S1: title. A supporting video article is available at doi: link.}

\authorcontributions{Conceptualization, J.Z., W.S., C.Y.; Data creation, J.Z.; Funding acquisition, C.Y. and T.C.; Methodology, J.Z.; Project administration, W.S.; Software, J.Z.; Supervision, W.S. and C.Y.; Validation, W.S. and C.Y.; Writing---original draft, J.Z. and C.Y.; Writing---review \& editing, C.Y. All authors have read and agreed to the published version of the manuscript.}

\funding{This work was supported by the National Key Research and Development Program of China (Grant Number: 2019YFB1310900), the National Natural Science Foundation of China (Grant Number: 62073229), and Jiangsu Policy Guidance Program (International Science and Technology Cooperation) The Belt and Road Initiative Innovative Cooperation Projects (Grant Number: BZ2021016), the Research Fund of Horizon Robotics, and The Natural Science Foundation of the Jiangsu Higher Education Institutions of China (Grant Number: 22KJB520008).}

\institutionalreview{Not applicable}

\informedconsent{Not applicable}

\dataavailability{The research uses the KITTI (https://www.cvlibs.net/datasets/kitti) and the nuScenes (https://www.nuscenes.org) datasets.}


\conflictsofinterest{The authors declare no conflict of interest.} 





\begin{adjustwidth}{-\extralength}{0cm}
\reftitle{References}

\end{adjustwidth}




\end{document}